\documentclass[runningheads,a4paper]{llncs}

\usepackage{amssymb}
\usepackage{graphicx}

\usepackage{url}
\urldef{\mailsa}\path|{alfred.hofmann, ursula.barth, ingrid.haas, frank.holzwarth,|
\urldef{\mailsb}\path|anna.kramer, leonie.kunz, christine.reiss, nicole.sator,|
\urldef{\mailsc}\path|erika.siebert-cole, peter.strasser, lncs}@springer.com|    

\usepackage[utf8]{inputenc} 
\usepackage[T1]{fontenc}    
\usepackage{hyperref}       
\usepackage{url}            
\usepackage{booktabs}       
\usepackage{amsfonts}       
\usepackage{nicefrac}       
\usepackage{microtype}      
\usepackage[font=small,skip=3.5pt]{caption}

\usepackage{amsmath,amssymb} 
\usepackage{color}
\usepackage{amsmath}
\usepackage{amssymb}
\usepackage{algorithm}
\usepackage{algorithmicx}
\usepackage{algpseudocode}
\usepackage{algpascal}

\usepackage{mathrsfs}
\usepackage{soul}
\usepackage{multirow}
\usepackage{comment}
\usepackage{xfrac}
\usepackage{rotating}
\usepackage{hhline}
\usepackage[table]{xcolor}
\usepackage{mathtools}
\usepackage{wrapfig}
\hypersetup{
    colorlinks=true,
    linkcolor=blue,
    filecolor=blue,      
    urlcolor=blue,
    citecolor=blue,
}

\usepackage{subcaption}

\usepackage{pgfplots}
\usepackage{pgfplotstable}
\usepackage{filecontents,pgfplots}
\usepackage{tikz}
\usetikzlibrary{arrows,calc,shapes,snakes,positioning,matrix,arrows,decorations.pathmorphing,decorations.text}
\usepackage{sci}

\usepackage{graphicx}

\usepackage{caption}

\usepackage{indentfirst}

\usepackage{multicol}
\usepackage{arydshln}
\usepackage{multirow}
\usepackage{hyperref}
\hypersetup{
    colorlinks=true,
    linkcolor=blue,
    filecolor=blue,      
    urlcolor=blue,
}
\usepackage{subcaption}
\usepackage{float}
\usepackage{amsmath}

\newcommand\blfootnote[1]
{%
\begingroup
\vspace{-5pt}
\renewcommand\thefootnote{}
\footnotetext{#1}%
\endgroup
}

\begin{document}
\mainmatter  

\title{End-To-End Alzheimer's Disease Diagnosis\\and Biomarker Identification}

\titlerunning{End-To-End Alzheimer's Disease Diagnosis}
\authorrunning{Esmaeilzadeh et al.}  
 
\author{Soheil Esmaeilzadeh$^1$ \and Dimitrios Ioannis Belivanis$^1$, \\ Kilian M. Pohl$^2$ \and Ehsan Adeli$^1$}
\institute{$^1$Stanford University \quad $^2$SRI International \\
\email{\{soes, dbelivan, eadeli\}@stanford.edu \quad kilian.pohl@sri.com}
}

\toctitle{End-To-End Alzheimer's Disease Diagnosis and Biomarker Identification}
\tocauthor{Soheil Esmaeilzadeh, Dimitrios Ioannis Belivanis, Kilian M. Pohl, Ehsan Adeli}
\maketitle

\begin{abstract} 
 
As shown in computer vision, the power of deep learning lies in automatically learning relevant and powerful features for any perdition task, which is made possible through end-to-end architectures. However, deep learning approaches applied for classifying medical images do not adhere to this architecture as they rely on several pre- and post-processing steps. This shortcoming can be explained by the relatively small number of available labeled subjects, the high dimensionality of neuroimaging data, and difficulties in interpreting the results of deep learning methods. In this paper, we propose a simple 3D Convolutional Neural Networks and exploit its model parameters to tailor the end-to-end architecture for the diagnosis of Alzheimer's disease (AD). Our model can diagnose AD with an accuracy of 94.1\% on the popular ADNI dataset using only MRI data, which outperforms the previous state-of-the-art. Based on the learned model, we identify the disease biomarkers, the results of which were in accordance with the literature. We further transfer the learned model to diagnose mild cognitive impairment (MCI), the prodromal stage of AD, which yield better results compared to other methods. 

\end{abstract}

\blfootnote{This work was supported in part by NIH grants {AA005965}, {AA017168}, and {MH11340-02}, and benefited from the NIH Cloud Credits Model Pilot. The authors also thank the investigators within ADNI ({\url{http://adni.loni.ucla.edu}}).}

\section{Introduction}
Alzheimer's disease is one of the most growing health issues, which devastated many lives, and the number of people with Alzheimer's dementia is predicted to be doubled within the next 20 years in the United States \cite{AlzheimersAssociation2017}. However, the basic understanding of the causes and mechanisms of the disease are yet to be explored. Currently, diagnosis is mainly performed by studying the individual's behavioral observations and medical history. Magnetic Resonance Imaging (MRI) is also used to analyze the brain morphometric patterns for identifying disease-specific imaging biomarkers. 

In recent years, numerous methods are introduced exploiting MRI data for distinguishing Alzheimer's Disease (AD) and its prodromal dementia stage, Mild Cognitive Impairment (MCI), from normal controls (NC). These approaches can be categorized in four main categories: Voxel-based methods \cite{Liu2018}, methods based on Regions-of-Interest (ROI) \cite{Kloppel2008,laakso1996hippocampal}, patch-based methods \cite{Liu2017}, and approaches that leverage features from whole-image-levels (\ie without considering local structures within the MRIs) \cite{Wolz2012}. The voxel-based approaches are prone to overfitting \cite{litjens2017survey} (due to high dimensionality input image) while ROI-based methods are confined to a coarse-scale limited number of ROIs \cite{litjens2017survey} that may neglect crucial fine-scaled information secluded within or across different regions of the brain. Patch-based approaches often ignore global brain representations and focus solely on fixed-size rectangular (or cubic) image patches. In contrast, whole-image approaches cannot identify the subtle changes in fine brain structures. Leveraging a trade-off between the global and local representations can, therefore, contribute to a better understanding of the disease, while not overemphasizing one aspect.

With the recent developments of deep learning and Convolutional Neural Network (CNN) algorithms in computer vision studies, many such methods are developed for medical imaging applications. However, the majority of such previous works mainly focused on segmentation, registration, landmark or lesion detection \cite{litjens2017survey}. For disease diagnosis, researchers have tried two-dimensional (2D) or three-dimensional (3D) patch-based models to train deep networks that diagnose diseases to a patch-level rather than subject-level. Only a few end-to-end deep learning methods (leveraging local and global MRI cues) are developed for the classification of neuroimages into different diagnostic groups \cite{Liu2017,litjens2017survey}, despite the power of deep learning owes to automatic feature learning made possible through end-to-end models. Not developing end-to-end models were mainly due to several limitations including: (1) not having enough labeled subjects in the datasets to train fully end-to-end models; (2) brain MRIs are 3D structures with high dimensionalities, which cause large computational costs; and (3) difficulties in interpretability of the results of end-to-end deep learning techniques from a neuroscience point-of-view. To resolve these challenges, instead of replicating standard deep learning architectures used in the computer vision domain, one requires explicit considerations and architectural designs. We conduct several experiments and tailor our architecture (through exploiting its numerous hyperparameters and architectural considerations) for classification of 3D MR images.

In this paper, we build a 3D Convolutional Neural Network (3D-CNN) and provide a simple method to interpret different regions of the brain and their association with the disease to identify AD biomarkers. Our method uses minimal preprocessing of MRIs (imposing minimum preprocessing artifacts) and utilizes a simple data augmentation strategy of downsampled MR images for training purposes. Unlike the vast majority of previous works, the proposed framework, thus, uses a voxel-based 3D-CNN to account for all voxels in the brain and capture the subtle \textit{local} brain details in addition to better pronounced \textit{global} specifics of MRIs. Using this detailed voxel-based representation of MRIs, we eliminate any a priori judgments for choosing ROIs or patches and take into account the whole brain. To avoid overfitting potentially caused by the large dimension of images, we carefully design our training model's architecture in a systematic way (not using standard computer vision architectures). We, then, propose a simple method to identify the MRI biomarkers of the disease by observing how confidently different regions of the brain contribute to the correct classification of the subjects. Finally, we propose a learning transfer strategy for MCI classification alongside the other two classes, in a three-class classification setting (AD, MCI, NC). Experiments on ADNI-1 dataset show superior results of our model compared to several baseline and prior works.

\section{Dataset and Preprocessing}
In this study, the public Alzheimer's Disease Neuroimaging Initiative-1 (ADNI-1) \cite{jack2008alzheimer} dataset is used, with all subjects having baseline brain T1-weighted structural MRI scans. The demographic information of the studied subjects is reported in Table \ref{tab1}. According to clinical criteria, such as Mini-Mental State Examination (MMSE) scores and Clinical Dementia Rating (CDR) (see \url{http://adni.loni.usc.edu}), subjects were diagnosed with AD or MCI conditions. There is a total of 841 subjects with baseline scans in the dataset, including 200 AD, 230 NC, and 411 MCI.
%
%
Fig. \ref{f1_2} shows the age distribution of different classes. Almost half of the subjects in each male/female category are in the MCI stage. Note that this stage is quite difficult to classify (from NC or AD) as it is a transition state and has similarities with both other classes. As can be seen, subjects are distributed proportionally similar across the three classes with respect to their age. Besides, both male and female groups have approximately similar portions of patients in each of the classes. Although the three classes are similar with respect to both age and gender distributions, we consider these two factors as input features to the model, as they can be confounding factors in MRI studies \cite{adeli2018}. 

%

\begin{table}[t]
\begin{minipage}[b]{.5\textwidth }%
\centering
\caption{ADNI-1 subjects demographic information.} \label{tab1}
{
\renewcommand{\arraystretch}{1.45}
\resizebox{0.98\textwidth}{!} {
\begin{tabular}{ccccccccc}
\hline
\multirow{2}{*}{\rotatebox[origin=c]{90}{\bf Class}} & \multirow{2}{*}{\rotatebox[origin=c]{90}{\bf Sex}}  &\multirow{2}{*}{\rotatebox[origin=c]{90}{\bf Count}}  & \multicolumn{6}{c}{\bf Age} \\ \cline{4-9}
     &          &  & mean$\pm$std & min & 25\%  & 50\%  & 75\%  & max \\
\hline \hline
\multirow{2}{*}{\textbf{AD}} & M & 97  & 75.0$\pm$7.9 & 55.2 & 70.8  & 75.3 & 80.4 & 91.0 \\
&  F  & 103 & 76.1$\pm$7.4 & 56.5 & 71.1 & 77.0  & 82.3 & 87.9                        \\
\hline
\multirow{2}{*}{\textbf{MCI}} & M  & 265 & 75.4$\pm$7.3 & 54.6 & 71.0 & 75.4  & 80.7 & 89.8 \\
&   F        & 146 & 73.6$\pm$7.5 & 55.2 & 69.1 & 74.3 & 79.7 & 86.2 \\
\hline
\multirow{2}{*}{\textbf{NC}} & M & 112 & 76.1$\pm$4.7 & 62.2 & 72.5 & 75.8 & 78.5 & 89.7 \\
& F  & 118 & 75.8$\pm$5.2 & 60.0 & 72.1 & 75.6  & 79.1 & 87.7                              \\
\hline
\end{tabular}
}
}
\hrule height 0pt
\end{minipage}%
~~
\begin{minipage}[b]{.47\textwidth}
\centering
\includegraphics[width=0.98\linewidth,trim={0 0.9cm 0 0},clip]{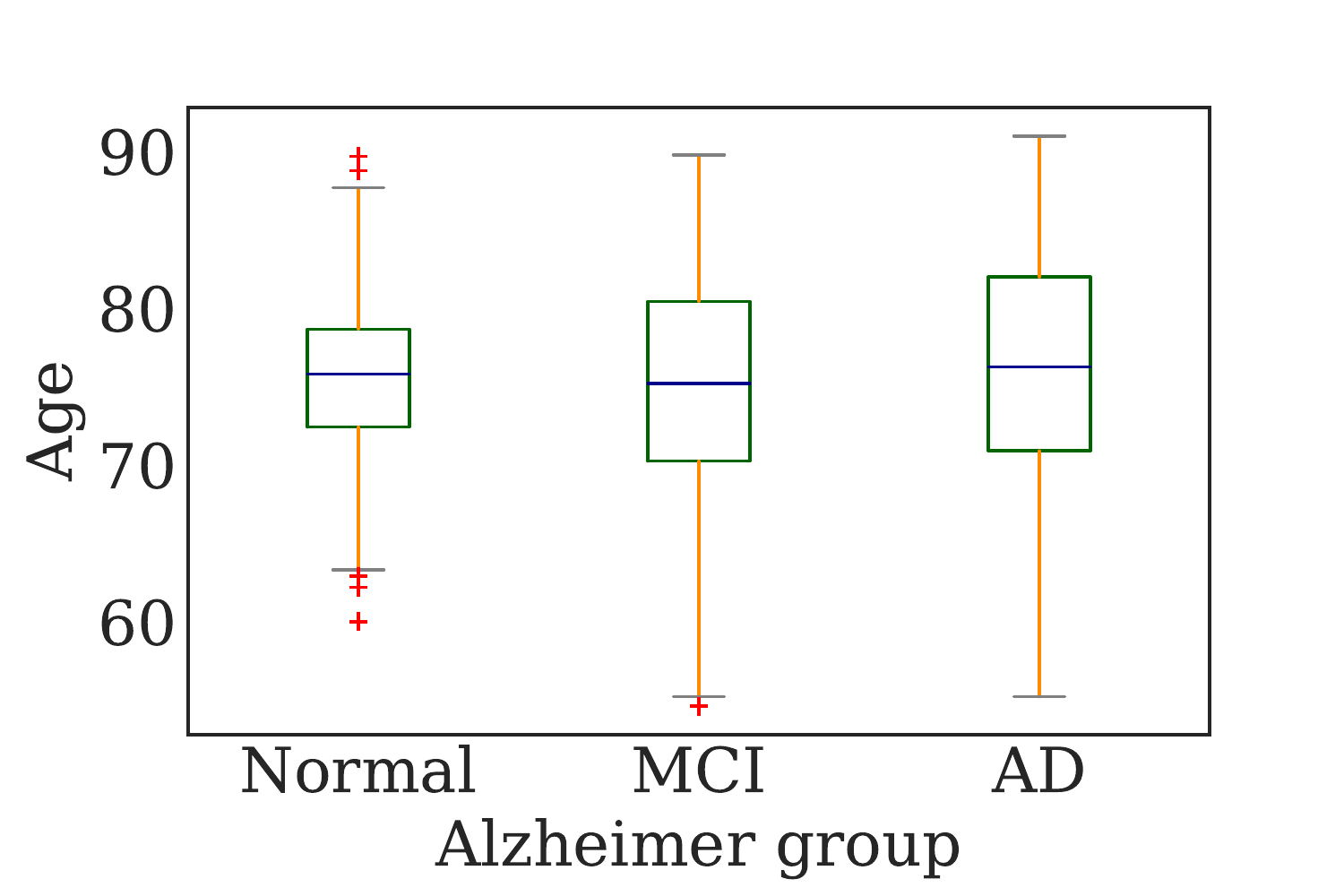}
\captionof{figure}{\strut Age distributions across groups. \label{f1_2}}
\end{minipage}
\end{table}

As a simple preprocessing step, the MR images of all subjects are skull-stripped, which includes removal of non-cerebral tissues like skull, scalp, and dura from brain images. To this end, we use the Brain Extraction Technique (BET) proposed in \cite{Smith2002}. This step reduces the size of images by a factor of two and hence slashes the amount of computational time spent for training the model.

\section{3D-CNN Training and Evaluation} \label{sec_Train_eval}

\begin{figure}[t]
\centering
\includegraphics[width=\linewidth]{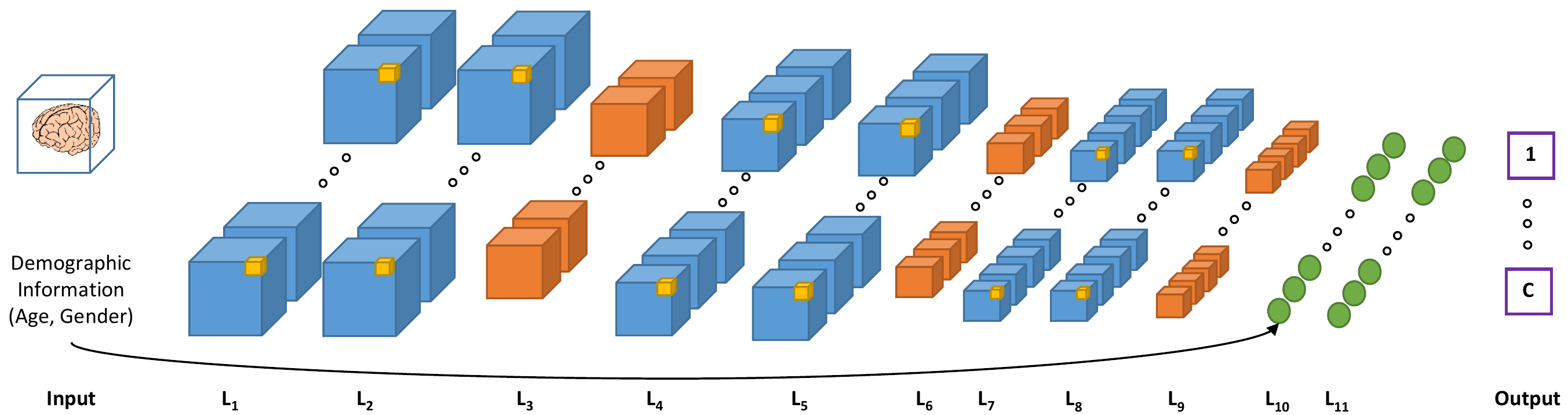}
\caption{3D-CNN architecture used in this paper. The blue cubes ($\text{L}_1$, $\text{L}_2$, $\text{L}_4$, $\text{L}_5$, $\text{L}_7$, and $\text{L}_8$) are convolutional layers; Orange cubes ($\text{L}_3$, $\text{L}_6$, and $\text{L}_9$) are max-pooling layers; and the last two layers are fully connected (FC) layers.}
\label{fig_CNNNetwork}
\end{figure}

\noindent\textbf{Architecture:} For our end-to-end classification task, we build a three-dimensional Convolutional Neural Network (3D-CNN) using the TensorFlow framework. To evaluate the performance and to avoid overfitting, we consider two architectures: a complex architecture, as shown in Fig. \ref{fig_CNNNetwork}, and a simplified version (with less number of filters, one less FC layer, and removing one Convolution (Conv.) layer at each stage). The complex architecture has $\mathcal{O}(10^5)$ trainable parameters, and the simple one has $\mathcal{O}(10^4)$ parameters. The fewer number of parameters helps the network avoid overfitting on a limited number of subjects.

The input MR images are re-sized to 116$\times$130$\times$83 voxels. The first batch of Conv. layers (L$_{1,2}$) have $3^3 \times 32$ filter and the second (L$_{4,5}$) and the third (L$_{7,8}$) $3^3 \times 64$ and $3^3 \times 128$, respectively. The max-pooling layers (L$_3$, L$_6$, and L$_9$) are with sizes $2^3$, $3^3$, and $4^3$, respectively. The fully connected (FC) layers have 512 (for L$_{10}$) and 256 (for L$_{11}$) nodes. The demographic variables of the subjects (age and gender) are added as two additional features in the first FC layer. We use a \textit{rectified linear unit (ReLU)} as the activation function, and a cross-entropy cost function as the loss, which is minimized with the \textit{Adam} optimizer. To optimize the architecture parameters and improve the trained model, we experiment by adding drop-out (D/O) and $\ell_2$-regularization (Reg). Therefore, several hyperparameters are introduced to experiment on, including the $\beta$ coefficient of the $\ell_2$-regularization, the drop-out probability, and the size of input training batches, in addition to the learning rate, number of filters in the convolutional layers, and the number of neurons in the FC layers. 

\noindent\textbf{Data Augmentation:} To train the model, we augment the data by flipping all subjects such that left and right hemispheres are swapped. This is a common strategy for data augmentation in the medical imaging as the neuroscientific studies suggest that the neurodegenerative disease (such as AD) impair the brain bilaterally \cite{AlzheimersAssociation2017}.

\noindent\textbf{Training Strategy:}
As can be seen in Fig. \ref{fig_CNNNetwork}, the output layer defines $c$ different classes. To better model the disease and identify its biomarkers, we first train the model on two classes ($c=2$), \ie AD and NC. After training the classifier with two classes, we add a third class (\ie MCI) and fine-tune the weights to now classify the input into three categories. This is simply possible as we use a cross-entropy loss in the last layer of the network, which can be easily extended for multi-class cases. This fine-tuning strategy is actually conducting a transfer learning from the domain of the two-class learned model to the three-class case. We show in our experiments that, in the presence of limited sets of training data such as medical imaging applications, this transfer learning strategy leads to better results compared to training the three-class model from scratch. It is important to note that MCI is the intermediate stage between the cognitive decline of normal aging and the more pronounced decline of dementia (to some extent between AD and NC), and hence, first learning to separate AD from NC identifies the differences between the two classes. Then, adding the third class and fine-tuning the network transfers the learned knowledge to classify the middle condition, not jeopardizing the performance of AD Diagnosis. 


\noindent\textbf{Evaluation:}
We use the classification accuracy (Acc), $F_2$-score, precision (Pre) and recall (Rec) for evaluating the models. Having true positive, true negative, false positive, and false negative denoted by $TP$, $TN$, $FP$, and $FN$, respectively, precision and recall are computed as 
%
  $\text{Pre} =\sfrac{TP}{(TP+FP)}$, $\text{Rec} =\sfrac{TP}{(TP+FN)}$.
and then the $F_2$-score is defined by weighing recall higher than precision (\ie placing more emphasis on false negatives, which is important for disease diagnosis):
  $F_2 = \sfrac{(5\times \text{Pre}  \times \text{Rec})}{(4\times \text{Pre}+\text{Rec})}$.
\section{Experiment Results}
%
%
To evaluate the model, at each iteration of 10-fold cross-validation, we randomly split the dataset into three sets of training (80\%), validation (10\%), and testing (10\%). Starting from the training model shown in Fig. \ref{fig_CNNNetwork} (the \textit{complex} architecture), we simplified the network, as described before, to avoid early overfitting. Besides, we investigated the effect of $\ell_2$-regularization of kernels and biases in the Conv. layers, as well as the FC layers with the regularization hyperparameter searched in the set \{0.01, 0.05, 0.1, 0.5, 1.0\}. Regularization coefficient 0.5 for the kernels and 1.0 for the biases are found to result in the best validation $F_2$-score. We also tested the drop-out strategy in the last two FC layers in the training process, controlling the drop-out extent by the value of keep-rate. We tested regularized simple and complex model architectures with different keep-rate values for the FC layers ranging from 0.15 to 0.85 and found that keep-rates of 0.15 and 0.25 for the first and second FC layers lead to the best validation-set accuracy in the complex model and keep-rate of 0.4 gives the best validation-set accuracy in the simple model.


\begin{table}[!t]
\centering
\caption{Ablation tests: testing performance comparison of different models (last row is our model). The comparison includes the Accuracy (Acc), $F_2$ score, Precision (Pre), and Recall (Rec) of all methods (Reg: Regularization, D/O: Drop-Out, Aug: Augmentation).}
\label{tbl:res}
\resizebox{1\textwidth}{!} {
\begin{tabular}{lC{1cm}C{1cm}C{1cm}C{1cm}cC{1cm}C{1cm}C{1cm}C{1cm}}
\hline
\multirow{2}{*}{\bf Model} & \multicolumn{4}{c}{\bf Simple} & ~ & \multicolumn{4}{c}{\bf Complex}\\ \cline{2-5} \cline{7-10}
 & {\bf Acc\%}& {\bf F$_2$} & {\bf Pre} & {\bf Rec} & & {\bf Acc\%}& {\bf F$_2$} & {\bf Pre} & {\bf Rec} \\
\hline \hline
3D-CNN          & 68.7 & 0.71   & 0.68  & 0.72  &   & 66.5 & 0.69 & 0.67   & 0.70\\
3D-CNN+Reg      & 77.6  & 0.77   & 0.74  & 0.78  &   & 77.4 & 0.75 & 0.72   & 0.76\\
3D-CNN+Reg+D/O  &  83.1 & 0.811  & 0.78  & 0.82  &   & 79.7 & 0.82 & 0.79   & 0.84  \\
\hline
3D-CNN+Reg+D/O+Aug (Ours) & \textbf{94.1} & \textbf{0.93} &\textbf{0.92} & \textbf{0.94} & & {88.3} & {0.89} & 0.88 & 0.91\\
\hline
\end{tabular}}\label{tab_1}
\end{table}


\noindent\textbf{AD \vs NC Classification Results (Two-Class Case):} Table \ref{tbl:res} shows the results of our model on the testing set in comparison with respect to ablation tests (removing components from the model and monitor how the performance changes). To test the significance of the classification results, we test our models using a Fisher exact test, in which our simple and complex models led to a $p$-value of less than 0.001. This indicates that the classifiers are significantly better than chance. As it can be seen, augmenting the size of the dataset led to improvement in the testing $F_2$ score, increasing it by 12.2\% from its value of 81.1\% in the non-augmented case. Another interesting observation is that the simple network outperforms the complex one, as it is less prone to overfitting.

%
%
%
Fig. \ref{fig_train_dev_acc} shows the training and validations accuracy and loss function values with respect to the number of epochs for the best model (\ie the one with validation-set $F_2$ score of 0.933). The learning process is terminated when the accuracy of the training set reaches near 1.0. Furthermore, the drop in the loss function curve after a middle stage plateau, where it reaches a saddle point, can be attributed to the hyperparameter tuning inherent to the \textit{Adam} optimizer during the training process. The model converges to a steady optimum without overfitting to the training data and hence yields reliable testing accuracies. 

\begin{table}[t]
\begin{minipage}[b]{.45\textwidth }%
\centering
\caption{Comparisons with prior works for AD diagnosis.} \label{tab:comp}
{
\renewcommand{\arraystretch}{1.45}
\resizebox{0.95\textwidth}{!} {
\begin{tabular}{c c C{1cm} C{1cm} C{1cm}}
\hline
{\bf Method} & {\bf Modalities} & {\bf Acc\%}  & {\bf Sen}  & {\bf Spe}   \\ 
\hline
\cite{suk2014hierarchical} & MRI+PET & 85.7 & 0.99 & 0.54 \\
\cite{hosseini2016alzheimer} & MRI & 90.8 & N/A & N/A \\
\cite{Liu2018} & MRI & 91.1 & 0.88 & 0.93 \\
\cite{khajehnejad2017alzheimer} & MRI & 93.9 & 0.94 & 0.93 \\
Ours & MRI & \textbf{94.1} & 0.94 & 0.91 \\
\hline
\end{tabular}
}
}
\hrule height 0pt
\end{minipage}%
~~
\begin{minipage}[b]{.5\textwidth}
\centering
\includegraphics[width=\linewidth]{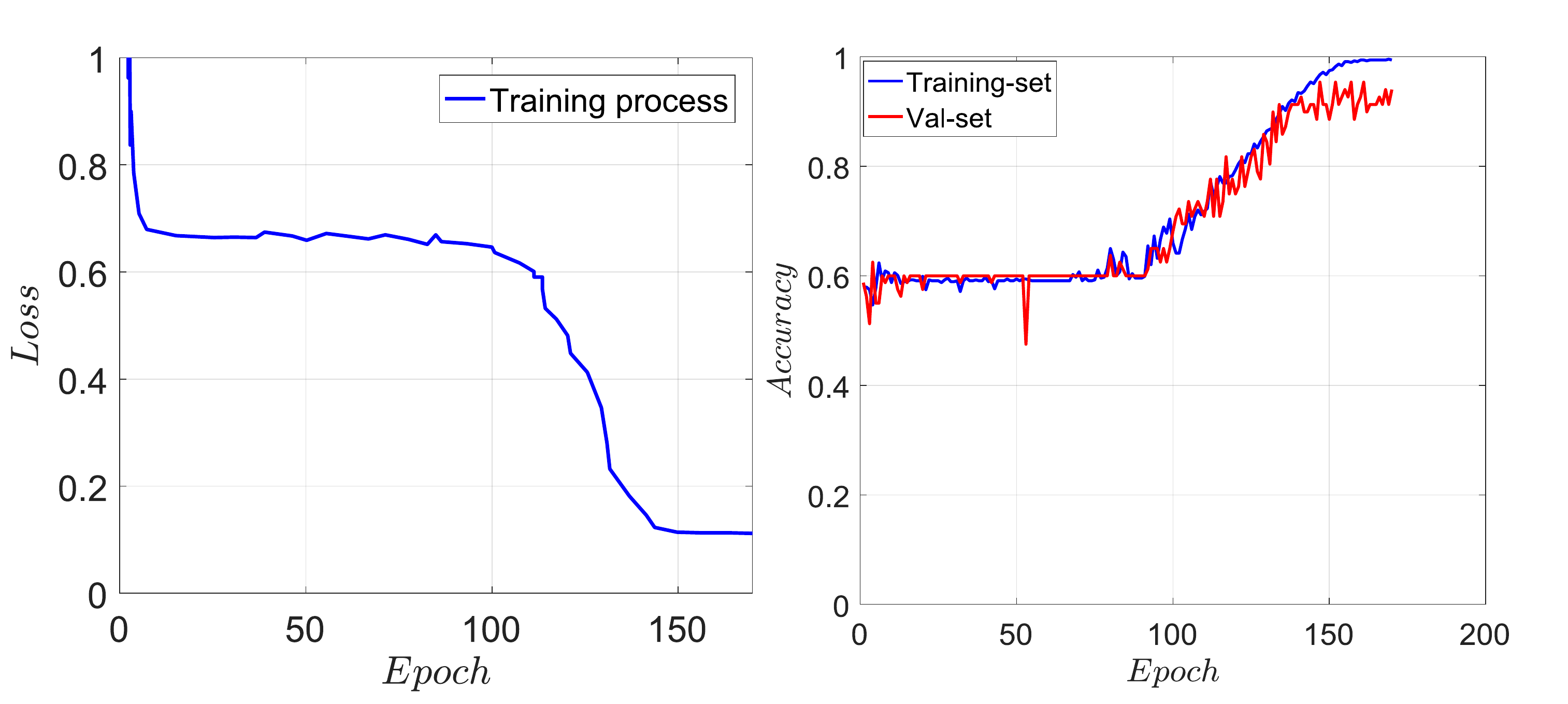}
\captionof{figure}{\strut (Left) training loss and (Right) training-validation accuracies with respect to the number of epochs for our 3D-CNN. \label{fig_train_dev_acc}}
\hrule height -10pt
\end{minipage}
\end{table}


%


\noindent\textbf{Comparisons with Prior Works:} Table \ref{tab:comp} compares the results of our AD \vs~NC classification with prior works in terms of accuracy, sensitivity (Sen), and Specificity (Spe) as reported in the respective references. Although the experimental setup in these references is slightly different, this table shows that our end-to-end classification model can classify the subjects more accurately. The improved accuracy can be attributed to the end-to-end manner of classifying the data, which helps to learn better features for the specific task of AD diagnosis and hence yield better results compared to other works.

\begin{figure}[t]
\centering
\includegraphics[trim={4cm 0 0 0},width=1\linewidth]{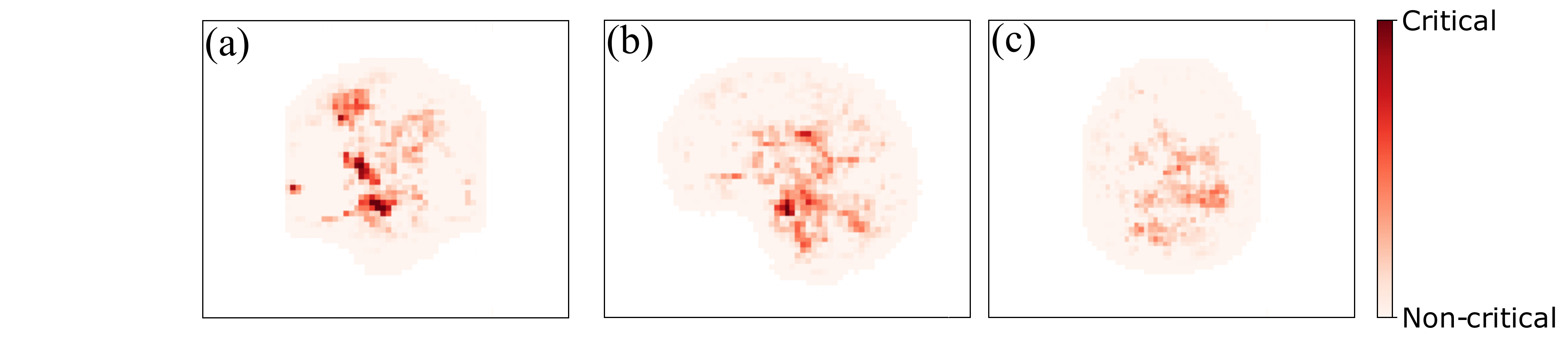}
\caption{Relative importance of different voxels associated with AD diagnosis.} 
\label{fig_Heat_map}
\end{figure}

\noindent\textbf{Identification of AD Biomarkers:}
To identify the regions of the brain that cause AD, we simply perform an image occlusion analysis on our best model (\ie 3D-CNN+Reg+D/O+Aug) by sliding a box of $1\times 1\times 1$ zero-valued voxels along the whole MR image of AD patients that were correctly labeled as AD by our trained model. The importance of each voxel, hence, can be characterized as the relative confidence of the samples being classified as AD. The resulting heat map is shown in Fig. \ref{fig_Heat_map}, in which the color map indicates the relative importance of each voxel. The red areas decrease the confidence of the model, suggesting that they are areas that are of critical importance in diagnosing AD. The red regions in Fig. \ref{fig_Heat_map} coincides with the hippocampus, amygdala, thalamus, and ventricles of the brain, which have been reported to be responsible for short-term memory and early stages of AD \cite{laakso1996hippocampal,AlzheimersAssociation2017,Liu2018,hosseini2016alzheimer}.

\begin{table}[!t]
\centering
\caption{Testing performance for three-class Alzheimer classification.}
\label{tbl:res01} 
\resizebox{\textwidth}{!} {
\begin{tabular}{lC{1.1cm}C{1cm}C{1cm}C{1cm}cC{1.1cm}C{1cm}C{1cm}C{1cm}}
\hline
\multirow{2}{*}{\bf Method} & \multicolumn{4}{c}{\bf Simple} & ~ & \multicolumn{4}{c}{\bf Complex}\\ \cline{2-5} \cline{7-10}
 & {\bf Acc\%}& {\bf F$_2$} & {\bf Pre} & {\bf Rec} & & {\bf Acc\%}& {\bf F$_2$} & {\bf Pre} & {\bf Rec} \\
\hline \hline
3D-CNN+D/O+Reg+with learning transfer      &   \textbf{61.1} &   \textbf{0.62} & \textbf{0.59}   & \textbf{0.63}  &   &  57.2 &   0.59 &  0.55   &  0.61\\
3D-CNN+D/O+Reg+w/o learning transfer        &  0.54 &  53.4 & 0.49   & 0.55  &   &  48.3 &   0.50 &  0.45   & 0.52 \\
\hline
\end{tabular}}
\end{table}

\noindent\textbf{Learning Transfer (Three-Class Classification):} We use the best model for the binary classification of AD \vs~NC in Table \ref{tab_1} and fine-tune it to develop a learning transfer strategy for classification MCI subjects. Doing so, we build a three-class classifier to classify NC \vs~MCI \vs~AD. To this end, the output layer of our model changes to $c=3$ instead of the previous $c=2$ classes. We keep the previously learned weights in the network and fine-tune the network by exposing it to the sample from the MCI class. Table \ref{tbl:res01} shows the results of training with learning transfer strategy, in comparison with the method that trains based on three classes from scratch. As it can be seen, our model results in 61.1\% accuracy, while if we train the model from scratch with all three classes, the model results in worse accuracies. This is due to the difficulty of the MCI class to distinguish from AD or NC. When training based on all three classes at once, the model gets stuck in local optima easier and overfit to the training data. On the other hand, the learning transfer strategy helps first learning the easy problem (\ie AD \vs~NC) and then transfer the knowledge to the domain of the harder class (\ie MCI). Interestingly, our three-class classification results are better than the results of other works for the three-class AD, MCI, and NC classification. For instance, Liu \etal~\cite{Liu2018} obtained a 51.8\% accuracy, compared to which, our results are better by a large margin (\ie 9.3\%). Again, this improvement can be attributed to the end-to-end design of our model and the learning transfer strategy.

%

\section{Conclusion}
In this paper, we developed a 3D-CNN model to diagnose Alzheimer's disease and its prodromal stage, MCI, using MR images. Our end-to-end model not only led to the best classification performance compared to other methods but also contributed to identifying relevant disease biomarkers. We found the \textit{hippocampus} region of the brain is critical in the diagnosis of AD. With an extensive hyperparameter tuning and exploiting the best model architecture for binary classification, we fine-tuned the resulting model for MCI diagnosis as well. An interesting finding of this work was that the simple architecture led to better testing results, compared to the other more complex architecture, as it is less prone to overfitting to the training data.
%
%

%

\bibliographystyle{splncs03}
{ \small
\bibliography{references}
}

%
%
%
%
%
%
\end{document}